\documentclass[letterpaper]{article}
\usepackage[preprint]{aaai2027}
\usepackage[hyphens]{url}
\usepackage{graphicx}
\usepackage{natbib}
\usepackage{caption}
\usepackage{booktabs}
\usepackage{amsfonts}
\usepackage{amsmath}
\usepackage{amssymb}
\usepackage{multirow}
\usepackage{array}

\newcommand{\nstar}{n^\star}

\title{What Does Chain-of-Thought Contribute at Probe Time? \\
Evidence for Local Co-Occurrence Activation}

\author{ Xiang Wang,
  Wei Wei
}

\affiliations{
  School of Computer Science and Technology\\
  Huazhong University of Science and Technology\\
  \{xiangwang, weiw\}@hust.edu.cn
}

\begin{document}
\maketitle

\begin{abstract}
Chain-of-thought (CoT) prompting enhances large language model performance, yet what drives these gains remains unclear. We study this question from a probe-time perspective: holding CoT rationales fixed, we test which textual properties matter for the final prediction. Across multiple datasets and model configurations, we find that randomizing the order of rationale sentences has little effect on accuracy, suggesting that the global order of reasoning steps is not the main source of the probe-time benefit. Moreover, even when the words in a rationale are randomly reordered, performance remains well above the no-rationale baseline, indicating that the rationale's words remain useful even without their original order. Restoring only short-range word order further improves performance and brings it substantially closer to full CoT. In most settings, much of this local-order gain is already obtained with three-word windows.
Control experiments rule out explicit answer copying, simple lexical cues, generic topical context, and general robustness to shuffling as the main explanations. Mechanistic analyses further show that short-window gains are largely formed in early-to-middle model layers, with answer-relevant evidence concentrated in local text spans. Together, these findings support a local co-occurrence activation (LCA) interpretation: the probe-time benefit of fixed rationales arises mainly from the words they contain and short-range word co-occurrences.
\end{abstract}

\section{Introduction}
Chain-of-thought (CoT) prompting has proven effective in improving language model accuracy \cite{wei2022chain,kojima2022large}. Prior work has largely focused on the generation side of CoT, studying how prompting strategies affect reasoning quality \cite{wang2023self,nye2021scratchpad,sprague2024analysis,qiao2023survey}, whether generated rationales faithfully reflect internal computation \cite{turpin2023unfaithful,lyu2023faithful,lin2024monitoring,matton2025walk,faithcot2025}, and what mechanisms underlie stepwise reasoning behavior \cite{olsson2022induction,feng2023mystery,cabannes2024iteration,chen2025howdoes}. Recent studies further suggest that CoT traces may serve as structural or distributional signals rather than faithful executable derivations \cite{wang2023towards,pfau2024dotbydot}. Together, these studies highlight a question: what properties of CoT actually drive its performance gains?

We study this question from a \emph{probe-time} perspective. Rather than examining how a rationale is generated, we hold the generated rationale fixed and study how its text contributes to the final prediction when used as context. This perspective separates the effect of the rationale text from the generation process, allowing us to perturb its structure in a controlled way and directly measure how these changes affect the probe-time benefit.

\begin{figure*}[t]
    \centering
    \includegraphics[width=\textwidth]{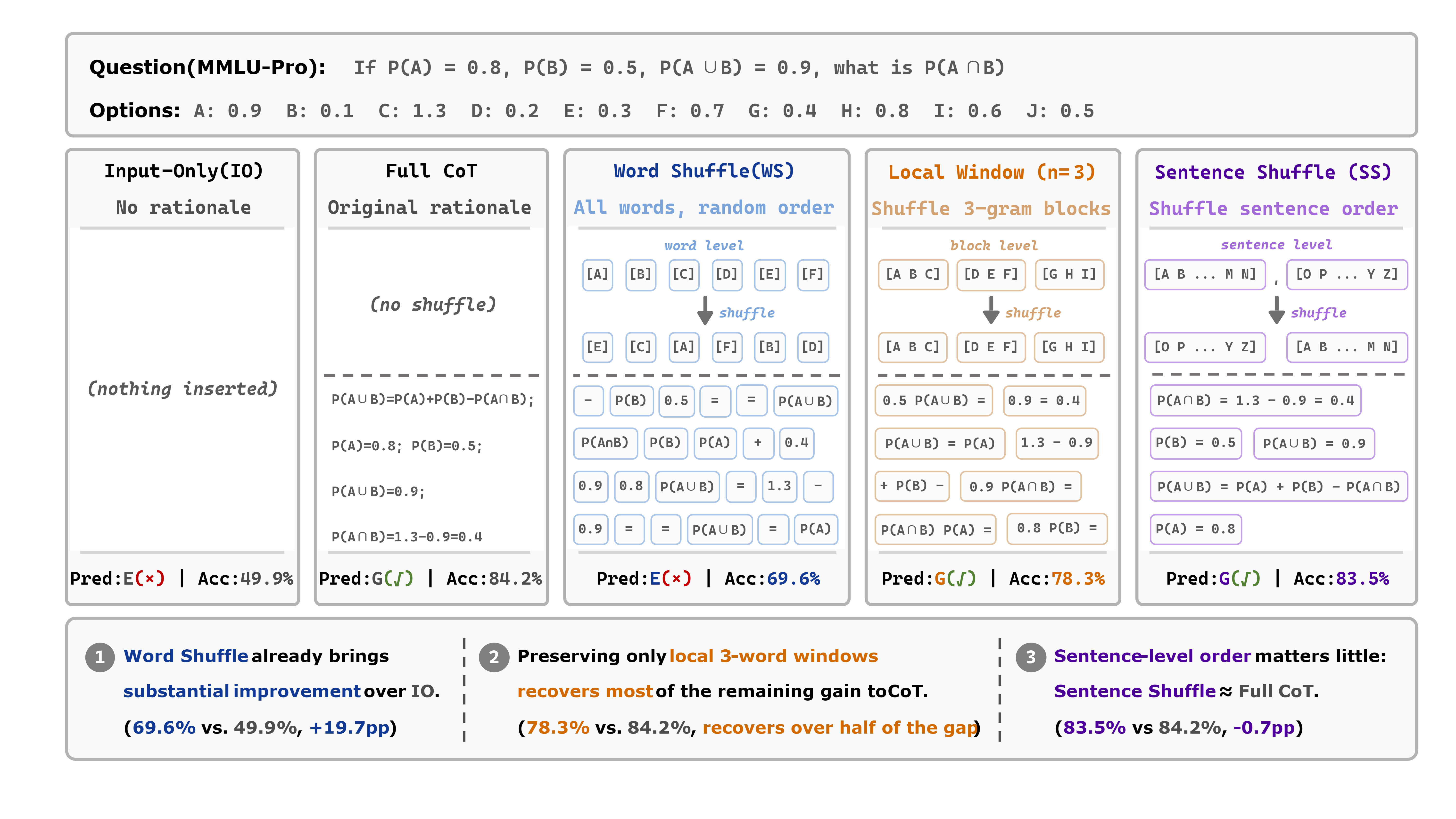}
    \caption{Motivating example of probe-time CoT perturbations. 
The top row shows one MMLU-Pro question. 
The five middle panels show IO, full CoT, word shuffle (WS), local window with $n{=}3$, and sentence shuffle (SS), including how the rationale is modified, the prediction on the example, and the pooled accuracy over 1,500 examples from Config~D. 
The bottom row highlights the main conclusions: word inventory gives a large gain over IO, short local windows recover much of the remaining gain, and sentence order matters little.
}
    \label{fig:motivating}
\end{figure*}

\begin{figure*}[t]
    \centering
    \includegraphics[width=\textwidth]{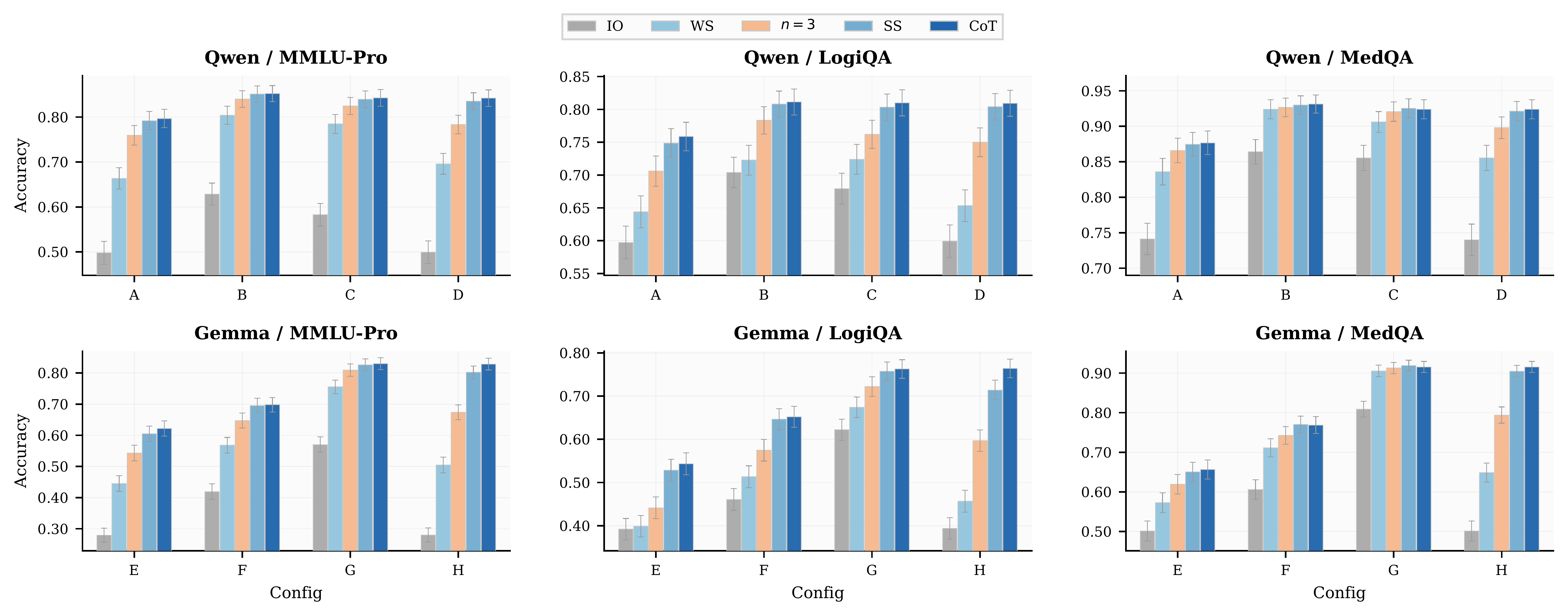}
    \caption{Probe-time accuracy under five conditions (IO, CoT, sentence-shuffle SS, word-shuffle WS, and $n{=}3$ blocks) across eight model configurations and three datasets. Error bars show 95\% binomial confidence-interval half-widths. Sentence-shuffled CoT closely tracks full CoT in every panel, while the word bag is substantially lower, and the $n{=}3$ condition already recovers much of the gap.}
    \label{fig:overview_bar}
\end{figure*}

\begin{figure*}[t]
    \centering
    \includegraphics[width=\textwidth]{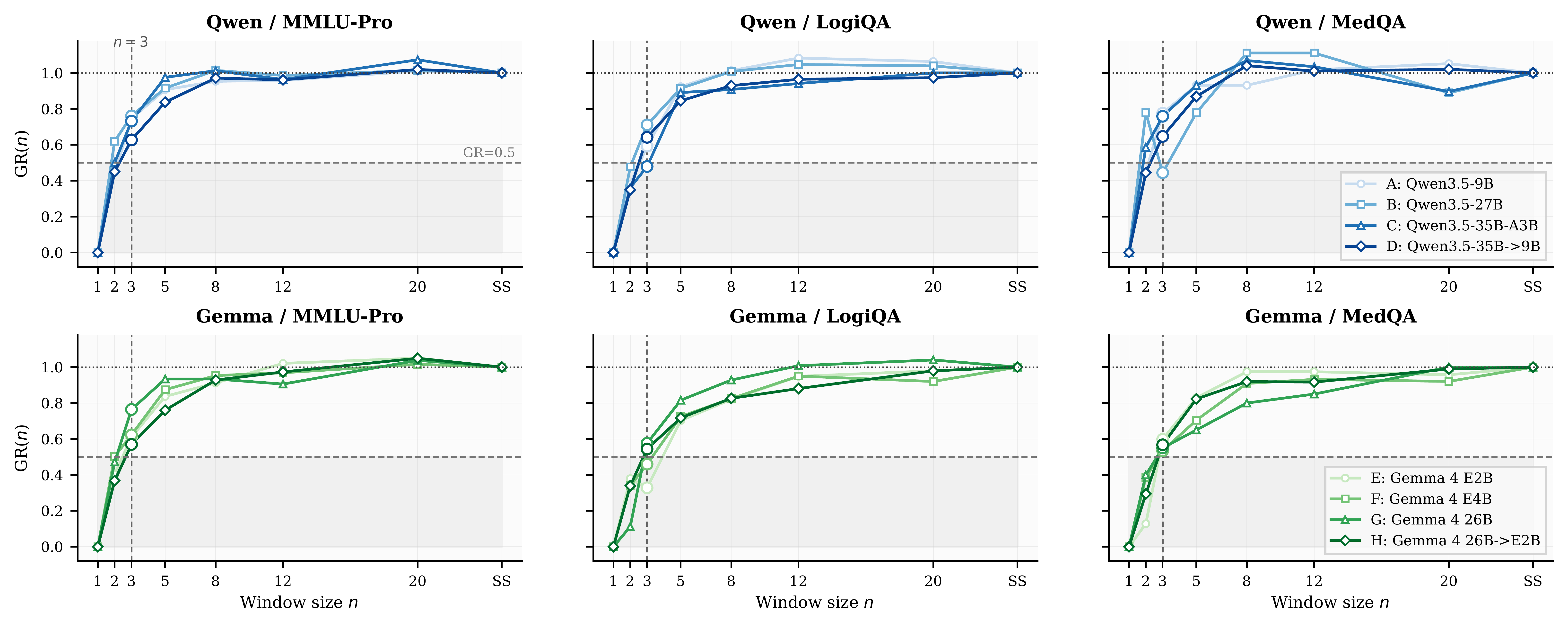}
    \caption{N-gram gap-recovery curves GR$(n)$ across eight configurations and three datasets, normalized between word shuffling (GR$=0$) and sentence shuffling (GR$=1$). Each line corresponds to one model configuration; the shaded region marks GR$<$0.5, the horizontal dashed line marks the half-recovery threshold, and the vertical dashed line/open markers highlight $n{=}3$.}
    \label{fig:main_ngram}
\end{figure*}

The exemplar in Figure~\ref{fig:motivating} illustrates our main observations. We begin by establishing two reference points: input-only prompting (IO) achieves 49.9\% accuracy, while full CoT reaches 84.2\%. Starting from these baselines, we progressively disrupt the structure of the rationale to examine where the probe-time benefit comes from. First, the global order of reasoning steps appears to matter little. Sentence shuffling (SS) preserves each sentence but randomizes sentence order, yet accuracy drops only slightly to 83.5\%. Second, the rationale's words remain useful even without their original order. Word shuffling (WS) randomizes all rationale words while preserving the same word inventory, and still achieves 69.6\%, well above IO. Finally, restoring only short local word order recovers much of the remaining benefit. We shuffle contiguous $n$-word windows while preserving word order within each window, with $n{=}1$ equivalent to WS. Preserving only three-word windows ($n{=}3$) raises accuracy to 78.3\%, recovering more than half of the improvement from WS to SS. Together, these observations suggest that the probe-time benefit depends strongly on the rationale's words and short local structure, but much less on the global progression of the reasoning chain.

To test the generality of our findings, we evaluate the same perturbations across three multiple-choice benchmarks and eight generator--probe configurations, where the same qualitative pattern consistently holds. We further extend the analysis to open-ended mathematical tasks, where the pattern persists, although slightly larger local windows are sometimes required. To rule out simpler explanations, we conduct controls for explicit answer copying, lexical answer cues, generic topical context, and general robustness to shuffling. Answer-stripping and tail-removal experiments show that the gains are not driven by explicit conclusions or evidence concentrated near the end of the rationale, while controls with topical reference text and perturbed question stems further indicate that the short-window effect is specific to rationale context. Mechanistic analyses provide complementary evidence: residual patching shows that short-window gains are largely formed in early-to-middle model layers, while text-span analysis shows that answer-relevant evidence is concentrated in local spans.

Together, these findings support a \emph{local co-occurrence activation} (LCA) interpretation. Under this view, the probe-time benefit of fixed rationales arises mainly from their word inventory and short-range word co-occurrences, rather than from the global order of reasoning steps.

Our contributions are:
(1) We introduce a probe-time protocol that holds CoT rationales fixed, isolating the effect of the rationale text from its generation process and enabling controlled perturbations of its structure.
(2) Across multiple datasets, model configurations, and task formats, we show that the global order of reasoning steps is not the main source of the probe-time benefit. The rationale's words remain useful even without their original order, while restoring only short-range word order yields substantial further gains. In most settings, three-word windows already provide much of this local-order gain.
(3) Through targeted controls and mechanistic analyses, we rule out explicit answer copying, simple lexical cues, generic topical context, and general robustness to shuffling as the main explanations. We further find that short-window gains emerge mainly in early-to-middle layers, with answer-relevant evidence concentrated in local text spans.

\section{Related Work}

\subsection{CoT Prompting and Faithfulness}

Chain-of-thought (CoT) prompting improves language-model accuracy by eliciting intermediate natural-language reasoning steps \citep{wei2022chain,kojima2022large}. Variants such as self-consistency and scratchpads can further improve performance on mathematical, symbolic, and knowledge-intensive tasks \citep{wang2023self,nye2021scratchpad,sprague2024analysis}, although the gains vary substantially across domains \citep{qiao2023survey}.

A parallel line of work studies whether generated rationales faithfully reflect a model's internal computation. Many studies show that language-model explanations can diverge from the true basis of a prediction, raising concerns about post-hoc rationalization and the reliability of CoT as an explanatory interface \citep{turpin2023unfaithful,lyu2023faithful,lin2024monitoring,matton2025walk,faithcot2025,barez2025explain,arcuschin2025wild}. For instance, \citet{lanham2023measuring} perturb self-generated rationales and show that models often ignore injected logical errors. Rather than studying the generation process or mechanistic faithfulness, we focus on a distinct probe-time question: once a rationale is fixed in context, which textual properties drive the probe model's final answer?

\subsection{Perturbation, Locality, and Prompt Structure}

Several studies suggest that CoT effectiveness may depend less on globally coherent derivations than on structural or distributional properties of rationale text. \citet{matton2025walk} show that symbolic structure can matter more than detailed semantic explanations in few-shot CoT prompts. \citet{wang2023towards} find that invalid reasoning steps can preserve much of the benefit of valid CoTs during generation, suggesting that relevance and structural consistency can matter even when exact logical correctness is disrupted. Other work shows that semantically meaningless filler words can substitute for CoT steps in some synthetic settings \citep{pfau2024dotbydot}.

Our work is also connected to analyses of locality, co-occurrence, and prompt redundancy. Robustness studies show that CoT behavior can be sensitive to paraphrastic, lexical, and adversarial perturbations \citep{xiang2024backdoor,xu2024preemptive,roh2025robustification,yue2025rethinking}. Mechanistic and theoretical work argues that CoT may externalize intermediate computation for bounded-depth transformers \citep{feng2023mystery}, induce iterative reasoning circuitry \citep{cabannes2024iteration,chen2025howdoes}, or benefit from local structure in the underlying learning problem \citep{prystawski2023why}. Prompt-compression work further suggests that long prompts contain substantial redundancy \citep{jiang2023llmlingua}. We build on these perspectives by isolating the probe-time contribution of fixed rationale text. We show that much of this benefit can be recovered from the rationale's word inventory and very short local word neighborhoods.

\section{Probe-Time Rationale Perturbations}

\subsection{Generator--probe protocol}

We study probe-time CoT in a controlled generator--probe setting. For each question, a generator model first produces a complete CoT rationale. We then insert this fixed rationale into the probe model's context as an assistant response, and ask the probe model to predict the final answer. This two-stage protocol gives accuracy close to Direct CoT in same-model settings, where the model generates its own rationale and answer in a single run. Complete results and prompt details are reported in the appendix under \emph{Direct CoT Baseline}.

Our evaluation spans three core datasets \citep{wang2024mmlu, liu2020logiqa, app11146421} and eight generator--probe configurations (A--H) across the Qwen3.5 \citep{qwen35blog} and Gemma 4 \citep{gemma} model families. To reduce seed-specific variation, we pool results across three random seeds (500 examples each; $n=1500$ per condition), and evaluate paired differences with two-sided McNemar's tests at $\alpha=0.05$. Comprehensive setup details are provided in the appendix under \emph{Experimental Details and Full Results}.

\subsection{N-gram perturbations}

To measure how much word order is needed beyond a rationale's word inventory, we introduce a family of controlled \emph{n}-gram perturbations. We use the word-shuffle (WS) condition as the lower reference point because it preserves the exact word selection but removes ordered text structure. This lets us test what happens when local order is gradually restored. Given the whitespace-tokenized word sequence of a rationale
\[
(w_1,w_2,\dots,w_T),
\]
we partition it into consecutive $n$-word blocks,
\[
(w_1,\dots,w_n),\ (w_{n+1},\dots,w_{2n}),\ \dots
\]
and randomly shuffle these blocks while preserving word order \emph{within} each block. When $n=1$, this procedure reduces to a global word bag. As $n$ increases, longer local word neighborhoods are preserved.

We measure this effect with a gap-recovery statistic, $\mathrm{GR}(n)$, which quantifies how much of the gap between word-shuffled (WS) and sentence-shuffled (SS) rationales is recovered at window size $n$:
\begin{equation}
\mathrm{GR}(n)=
\frac{
\mathrm{Acc}(n)-\mathrm{Acc}(\mathrm{WS})
}{
\mathrm{Acc}(\mathrm{SS})-\mathrm{Acc}(\mathrm{WS})
}.
\end{equation}

Here, $\mathrm{GR}(n)=0$ corresponds to the word-shuffle baseline, and $\mathrm{GR}(n)=1$ corresponds to sentence-shuffled performance. We use SS as the upper reference point because it preserves sentence-internal structure while removing the global order of reasoning steps. Thus, the $\mathrm{WS}\rightarrow\mathrm{SS}$ gap measures the additional benefit obtained by moving from a word bag to locally structured rationale text. Since SS performs close to full CoT in our experiments, this gap captures most of the recoverable structure beyond word inventory, without conflating it with the small difference between SS and full CoT.

We define the critical window size $n^\star$ as the smallest $n$ satisfying $\mathrm{GR}(n) \ge 0.5$. This means that window size $n$ has recovered at least half of the local-structure gain between WS and SS. If the probe model relied on long-range dependencies or full sentence-level coherence, $n^\star$ would be large. The \emph{n}-gram sweep tests whether this is the case.

\begin{figure*}[t]
    \centering
    \includegraphics[width=\textwidth]{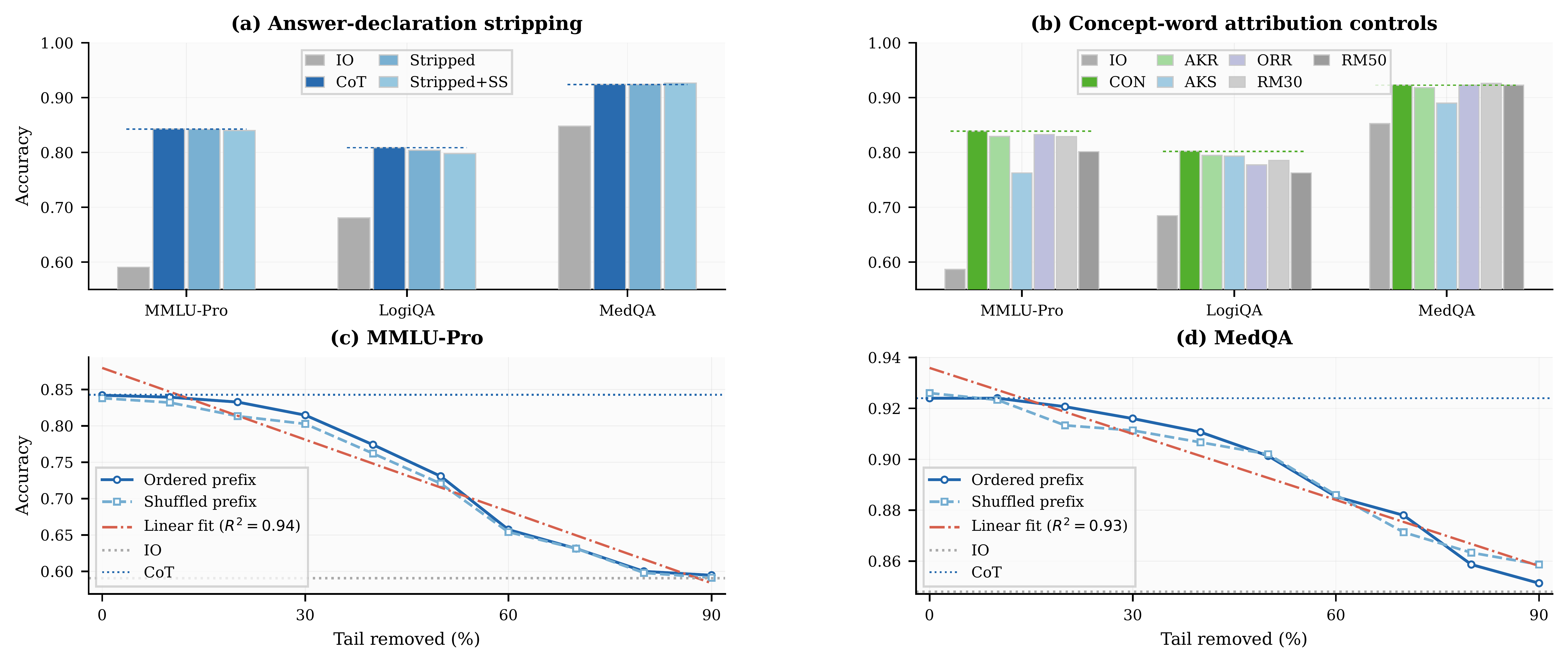}
    \caption{Answer-stripping (a), concept-word attribution (b), and tail-sweep (c,d) experiments. Removing explicit answer-declaration sentences produces little or no accuracy change (a); answer-keyword removal, keyword swapping, option-reference removal, and random masking cannot explain the full CoT gain (b); accuracy declines smoothly as larger rationale suffixes are removed, and the fitted linear trend highlights the near-linear tail-removal behavior (c,d).}
    \label{fig:strip_tail}
\end{figure*}

\subsection{Short windows recover much of the local-structure gap}

Figure~\ref{fig:overview_bar} summarizes the main perturbation results across all eight configurations. Full CoT improves over IO across datasets and models. Sentence shuffling (SS) stays close to full CoT, suggesting that the global order of reasoning steps is not the main source of the probe-time gain. Word shuffling (WS) scores clearly below SS but remains well above IO, showing that the rationale's word inventory already provides a substantial probe-time signal. The representative $n{=}3$ condition lies between these two cases and recovers a large part of the remaining gap with only short local context.

Figure~\ref{fig:main_ngram} shows the same transition through the full \emph{n}-gram sweep. Across datasets and model configurations, accuracy rises sharply once local order is restored beyond $n=1$. By $n{=}3$, most configurations have already recovered a large fraction of the $\mathrm{WS}\rightarrow\mathrm{SS}$ gap. This indicates that the probe model needs much less structure than a fully grammatical rationale sentence.

The recovery curves show a consistent pattern across both Qwen3.5 and Gemma~4 model families, and across both self-probing and cross-model probing settings. Although the absolute CoT gains vary by task, the critical window size $n^\star$ remains small in most settings. These results suggest that probe-time CoT benefits do not mainly depend on following a globally coherent reasoning chain. Instead, much of the recoverable structure beyond word inventory comes from immediate local word neighborhoods.

\section{Testing Alternative Explanations}

The short-window recovery pattern leaves several simpler explanations open. The probe model might copy explicit final-answer statements, follow local lexical cues, benefit from generic topical overlap, or simply be robust to shuffled text. We therefore use targeted controls to test these alternatives. For clarity, shorthand names for several perturbations, including AKR, AKS, ORR, and RM30/RM50, are defined in the appendix together with implementation details and additional controls.

\subsection{The Gain Is Not Driven by Answer Statements or Cues}

A natural hypothesis is that the probe model simply copies explicit answer statements near the end of the rationale, such as ``therefore the answer is C''. To test this, we automatically detect and remove all answer-declaration sentences before inserting the rationale into the probe context. Depending on the dataset, these sentences account for roughly 3--11\% of rationale words.

Figure~\ref{fig:strip_tail}(a) shows the resulting accuracy changes for Config~C. Removing answer declarations (Stripped CoT) causes little or no measurable degradation, and the same pattern holds after sentence shuffling (Stripped+SS). On MMLU-Pro and MedQA, the stripped and full CoT conditions are statistically indistinguishable under McNemar testing. On LogiQA, the difference is statistically detectable but only about half a percentage point. Thus, the main gain is not explained by explicit final-answer statements.

Answer stripping only removes direct answer declarations. A subtler possibility is that conclusion-oriented evidence is concentrated near the end of the rationale but does not match our answer-declaration templates. We therefore conduct a tail-sweep experiment, progressively removing rationale suffixes and measuring accuracy as more tail text is removed. If the gain were dominated by a final conclusion segment, accuracy should remain close to full CoT until that segment is removed, and then drop sharply. Instead, accuracy declines smoothly as larger suffixes are removed, with a fitted trend close to linear (Figure~\ref{fig:strip_tail}(c,d)). This suggests that useful probe-time signal is distributed across the rationale, rather than concentrated in a single final answer statement or summary.

We also test whether the probe relies on lexical answer cues rather than explicit answer declarations. We remove words copied from the correct answer option (AKR), replace such words with words from a randomly selected incorrect option (AKS), remove direct option references (ORR), and randomly mask 30\% or 50\% of concept words (RM30/RM50). Figure~\ref{fig:strip_tail}(b) summarizes the results. AKR causes only a small degradation. AKS produces a larger drop, but the probe follows the swapped incorrect answer only about 6--8\% of the time. Thus, the probe is sensitive to rationale content, but it does not simply output whichever answer string is made salient. Together, these controls show that explicit answer copying and simple answer-cue following are not the main explanations for the recovery pattern.

\begin{figure*}[t]
    \centering
    \includegraphics[width=\textwidth]{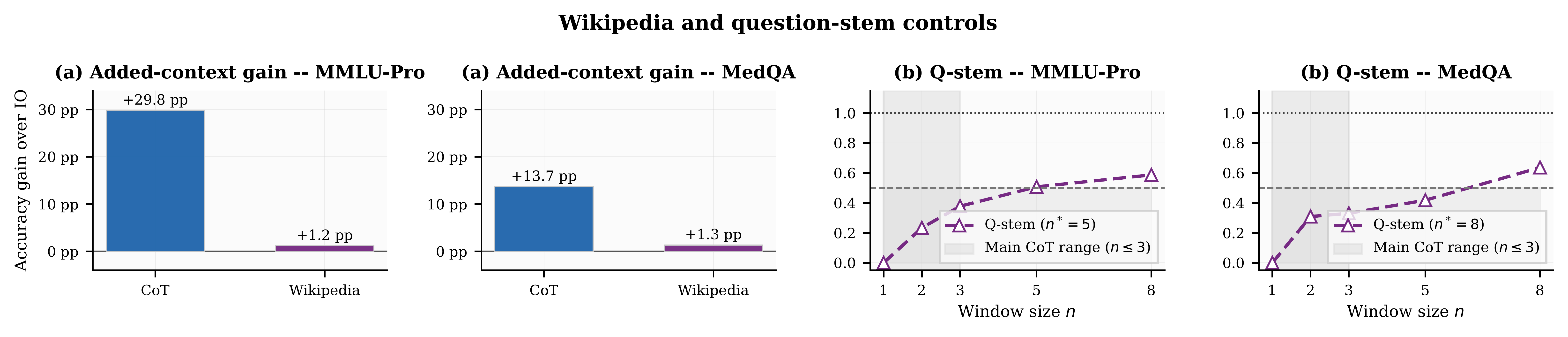}
    \caption{Wikipedia control (a) and question-stem control (b). Topic-matched Wikipedia text produces near-zero recovery, while ordinary question text requires substantially larger windows than CoT rationales.}
    \label{fig:wiki_qstem}
\end{figure*}

\begin{figure*}[t]
    \centering
    \includegraphics[width=\textwidth]{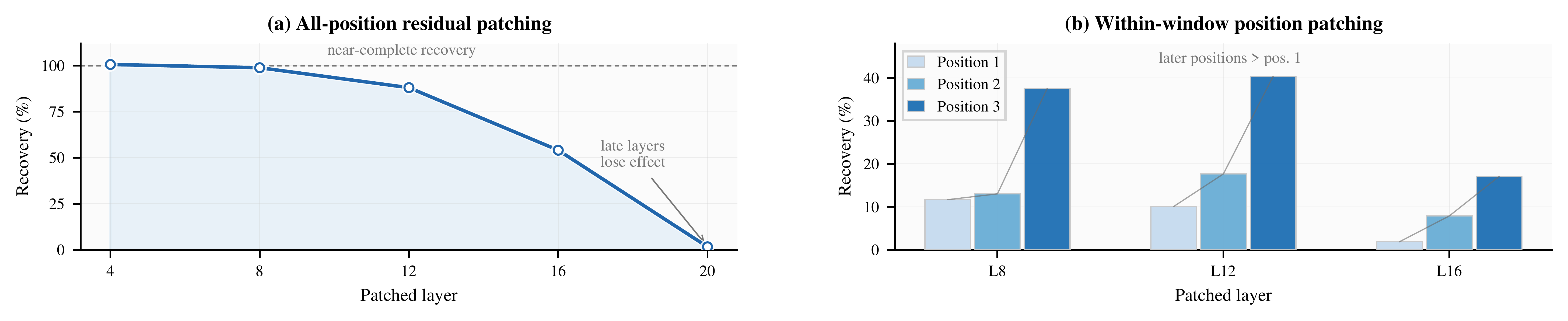}
    \caption{Causal patching results on MMLU-Pro Config~A. (a) All-position residual patching recovery from Layers 4 to 20. (b) Position-specific residual patching recovery for the first, second, and third word positions within each $n{=}3$ block at Layers 8, 12, and 16.}
\label{fig:causal_patching}
\end{figure*}

\begin{table*}[t]
\centering
\small
\setlength{\tabcolsep}{3.6pt}
\begin{tabular}{lr cccccccccc}
\toprule
\textbf{Dataset} & \multicolumn{1}{c}{\textbf{Config}} & IO & $n{=}1$ & $n{=}2$ & $n{=}3$ & $n{=}5$ & $n{=}8$ & $n{=}12$ & $n{=}20$ & SS & CoT \\
\midrule
\multirow{8}{*}{MMLU-Pro} & A ($\nstar=3$) & 49.8$\!\pm\!$2.5 & 66.3$\!\pm\!$2.4 & 72.1$\!\pm\!$2.3 & 75.9$\!\pm\!$2.2 & 78.0$\!\pm\!$2.1 & 78.6$\!\pm\!$2.1 & 78.7$\!\pm\!$2.1 & 79.3$\!\pm\!$2.0 & 79.2$\!\pm\!$2.1 & 79.7$\!\pm\!$2.0 \\
 & B ($\nstar=2$) & 62.9$\!\pm\!$2.4 & 80.4$\!\pm\!$2.0 & 83.3$\!\pm\!$1.9 & 84.0$\!\pm\!$1.9 & 84.7$\!\pm\!$1.8 & 85.2$\!\pm\!$1.8 & 85.1$\!\pm\!$1.8 & 85.2$\!\pm\!$1.8 & 85.1$\!\pm\!$1.8 & 85.2$\!\pm\!$1.8 \\
 & C ($\nstar=2$) & 58.3$\!\pm\!$2.5 & 78.5$\!\pm\!$2.1 & 81.2$\!\pm\!$2.0 & 82.5$\!\pm\!$1.9 & 83.8$\!\pm\!$1.9 & 84.0$\!\pm\!$1.9 & 83.7$\!\pm\!$1.9 & 84.3$\!\pm\!$1.8 & 83.9$\!\pm\!$1.9 & 84.3$\!\pm\!$1.8 \\
 & D ($\nstar=3$) & 49.9$\!\pm\!$2.5 & 69.6$\!\pm\!$2.3 & 75.9$\!\pm\!$2.2 & 78.3$\!\pm\!$2.1 & 81.3$\!\pm\!$2.0 & 83.1$\!\pm\!$1.9 & 83.0$\!\pm\!$1.9 & 83.8$\!\pm\!$1.9 & 83.5$\!\pm\!$1.9 & 84.2$\!\pm\!$1.8 \\
 & E ($\nstar=3$) & 27.9$\!\pm\!$2.3 & 44.5$\!\pm\!$2.5 & 50.9$\!\pm\!$2.5 & 54.3$\!\pm\!$2.5 & 57.9$\!\pm\!$2.5 & 59.1$\!\pm\!$2.5 & 60.8$\!\pm\!$2.5 & 61.3$\!\pm\!$2.5 & 60.5$\!\pm\!$2.5 & 62.2$\!\pm\!$2.5 \\
 & F ($\nstar=2$) & 41.9$\!\pm\!$2.5 & 56.8$\!\pm\!$2.5 & 63.2$\!\pm\!$2.4 & 64.7$\!\pm\!$2.4 & 67.9$\!\pm\!$2.4 & 68.9$\!\pm\!$2.3 & 69.1$\!\pm\!$2.3 & 69.7$\!\pm\!$2.3 & 69.5$\!\pm\!$2.3 & 69.8$\!\pm\!$2.3 \\
 & G ($\nstar=3$) & 57.0$\!\pm\!$2.5 & 75.5$\!\pm\!$2.2 & 78.9$\!\pm\!$2.1 & 80.9$\!\pm\!$2.0 & 82.1$\!\pm\!$1.9 & 82.1$\!\pm\!$1.9 & 81.9$\!\pm\!$1.9 & 82.9$\!\pm\!$1.9 & 82.6$\!\pm\!$1.9 & 83.0$\!\pm\!$1.9 \\
 & H ($\nstar=3$) & 28.0$\!\pm\!$2.3 & 50.5$\!\pm\!$2.5 & 61.4$\!\pm\!$2.5 & 67.4$\!\pm\!$2.4 & 73.1$\!\pm\!$2.2 & 78.1$\!\pm\!$2.1 & 79.4$\!\pm\!$2.0 & 81.7$\!\pm\!$2.0 & 80.2$\!\pm\!$2.0 & 82.9$\!\pm\!$1.9 \\
\midrule
\multirow{8}{*}{MedQA} & A ($\nstar=3$) & 74.1$\!\pm\!$2.2 & 83.6$\!\pm\!$1.9 & 85.3$\!\pm\!$1.8 & 86.6$\!\pm\!$1.7 & 87.2$\!\pm\!$1.7 & 87.2$\!\pm\!$1.7 & 87.5$\!\pm\!$1.7 & 87.7$\!\pm\!$1.7 & 87.5$\!\pm\!$1.7 & 87.7$\!\pm\!$1.7 \\
 & B ($\nstar=2$) & 86.4$\!\pm\!$1.7 & 92.4$\!\pm\!$1.3 & 92.9$\!\pm\!$1.3 & 92.7$\!\pm\!$1.3 & 92.9$\!\pm\!$1.3 & 93.1$\!\pm\!$1.3 & 93.1$\!\pm\!$1.3 & 92.9$\!\pm\!$1.3 & 93.0$\!\pm\!$1.3 & 93.1$\!\pm\!$1.3 \\
 & C ($\nstar=2$) & 85.5$\!\pm\!$1.8 & 90.6$\!\pm\!$1.5 & 91.7$\!\pm\!$1.4 & 92.1$\!\pm\!$1.4 & 92.4$\!\pm\!$1.3 & 92.7$\!\pm\!$1.3 & 92.6$\!\pm\!$1.3 & 92.3$\!\pm\!$1.3 & 92.5$\!\pm\!$1.3 & 92.4$\!\pm\!$1.3 \\
 & D ($\nstar=3$) & 74.0$\!\pm\!$2.2 & 85.5$\!\pm\!$1.8 & 88.5$\!\pm\!$1.6 & 89.8$\!\pm\!$1.5 & 91.3$\!\pm\!$1.4 & 92.4$\!\pm\!$1.3 & 92.2$\!\pm\!$1.4 & 92.3$\!\pm\!$1.4 & 92.1$\!\pm\!$1.4 & 92.4$\!\pm\!$1.3 \\
 & E ($\nstar=3$) & 50.1$\!\pm\!$2.5 & 57.3$\!\pm\!$2.5 & 58.3$\!\pm\!$2.5 & 61.9$\!\pm\!$2.5 & 63.7$\!\pm\!$2.4 & 64.9$\!\pm\!$2.4 & 64.9$\!\pm\!$2.4 & 64.7$\!\pm\!$2.4 & 65.1$\!\pm\!$2.4 & 65.7$\!\pm\!$2.4 \\
 & F ($\nstar=3$) & 60.6$\!\pm\!$2.5 & 71.1$\!\pm\!$2.3 & 73.4$\!\pm\!$2.2 & 74.3$\!\pm\!$2.2 & 75.3$\!\pm\!$2.2 & 76.5$\!\pm\!$2.1 & 76.6$\!\pm\!$2.1 & 76.5$\!\pm\!$2.1 & 77.0$\!\pm\!$2.1 & 76.9$\!\pm\!$2.1 \\
 & G ($\nstar=3$) & 80.9$\!\pm\!$2.0 & 90.5$\!\pm\!$1.5 & 91.1$\!\pm\!$1.4 & 91.3$\!\pm\!$1.4 & 91.4$\!\pm\!$1.4 & 91.6$\!\pm\!$1.4 & 91.7$\!\pm\!$1.4 & 91.9$\!\pm\!$1.4 & 91.9$\!\pm\!$1.4 & 91.5$\!\pm\!$1.4 \\
 & H ($\nstar=3$) & 50.1$\!\pm\!$2.5 & 64.9$\!\pm\!$2.4 & 72.4$\!\pm\!$2.3 & 79.4$\!\pm\!$2.0 & 85.9$\!\pm\!$1.8 & 88.4$\!\pm\!$1.6 & 88.3$\!\pm\!$1.6 & 90.2$\!\pm\!$1.5 & 90.5$\!\pm\!$1.5 & 91.5$\!\pm\!$1.4 \\
\midrule
\multirow{8}{*}{MATH-Hard} & A ($\nstar=3$) & 12.0$\!\pm\!$2.6 & 33.3$\!\pm\!$3.8 & 47.3$\!\pm\!$4.0 & 57.8$\!\pm\!$4.0 & 64.3$\!\pm\!$3.8 & 66.7$\!\pm\!$3.8 & 69.0$\!\pm\!$3.7 & 70.0$\!\pm\!$3.7 & 68.8$\!\pm\!$3.7 & 70.7$\!\pm\!$3.6 \\
 & B ($\nstar=3$) & 25.2$\!\pm\!$3.5 & 54.0$\!\pm\!$4.0 & 63.0$\!\pm\!$3.9 & 68.8$\!\pm\!$3.7 & 73.5$\!\pm\!$3.5 & 73.5$\!\pm\!$3.5 & 73.3$\!\pm\!$3.5 & 73.8$\!\pm\!$3.5 & 73.3$\!\pm\!$3.5 & 74.2$\!\pm\!$3.5 \\
 & C ($\nstar=3$) & 20.2$\!\pm\!$3.2 & 50.5$\!\pm\!$4.0 & 60.0$\!\pm\!$3.9 & 66.5$\!\pm\!$3.8 & 71.3$\!\pm\!$3.6 & 72.7$\!\pm\!$3.6 & 73.7$\!\pm\!$3.5 & 73.7$\!\pm\!$3.5 & 73.2$\!\pm\!$3.5 & 73.2$\!\pm\!$3.5 \\
 & D ($\nstar=3$) & 11.8$\!\pm\!$2.6 & 36.8$\!\pm\!$3.9 & 49.5$\!\pm\!$4.0 & 57.3$\!\pm\!$4.0 & 65.7$\!\pm\!$3.8 & 70.2$\!\pm\!$3.7 & 70.7$\!\pm\!$3.6 & 71.7$\!\pm\!$3.6 & 70.8$\!\pm\!$3.6 & 72.3$\!\pm\!$3.6 \\
 & E ($\nstar=5$) & 7.5$\!\pm\!$2.1 & 26.7$\!\pm\!$3.5 & 30.7$\!\pm\!$3.7 & 34.0$\!\pm\!$3.8 & 43.8$\!\pm\!$4.0 & 48.5$\!\pm\!$4.0 & 53.0$\!\pm\!$4.0 & 55.0$\!\pm\!$4.0 & 57.3$\!\pm\!$4.0 & 55.7$\!\pm\!$4.0 \\
 & F ($\nstar=5$) & 12.0$\!\pm\!$2.6 & 35.3$\!\pm\!$3.8 & 39.8$\!\pm\!$3.9 & 48.0$\!\pm\!$4.0 & 56.8$\!\pm\!$4.0 & 62.5$\!\pm\!$3.9 & 63.8$\!\pm\!$3.8 & 65.3$\!\pm\!$3.8 & 68.3$\!\pm\!$3.7 & 64.2$\!\pm\!$3.8 \\
 & G ($\nstar=5$) & 17.7$\!\pm\!$3.1 & 51.8$\!\pm\!$4.0 & 57.7$\!\pm\!$4.0 & 64.3$\!\pm\!$3.8 & 72.2$\!\pm\!$3.6 & 74.2$\!\pm\!$3.5 & 76.7$\!\pm\!$3.4 & 77.3$\!\pm\!$3.4 & 77.8$\!\pm\!$3.3 & 77.7$\!\pm\!$3.3 \\
 & H ($\nstar=5$) & 7.3$\!\pm\!$2.1 & 28.3$\!\pm\!$3.6 & 40.2$\!\pm\!$3.9 & 43.7$\!\pm\!$4.0 & 58.0$\!\pm\!$3.9 & 62.7$\!\pm\!$3.9 & 70.2$\!\pm\!$3.7 & 73.7$\!\pm\!$3.5 & 79.7$\!\pm\!$3.2 & 76.7$\!\pm\!$3.4 \\
\bottomrule
\end{tabular}
\caption{Accuracy (\%) across representative multiple-choice and open-ended settings. Each cell reports the point estimate with a 95\% binomial confidence-interval half-width.}
\label{tab:cross_ngram}
\end{table*}

\subsection{The Effect Is Specific to Rationale Context}

Another possibility is that the effect is not specific to CoT-style rationales. Any topical reference passage of similar length might provide enough lexical overlap to help the probe model. To test this, we inject BM25-retrieved Wikipedia passages matched to the original CoTs by topic and approximate length. Unlike rationale text, these passages produce near-zero gains. This shows that generic topical overlap does not reproduce the benefit of rationale text.

A related possibility is that language models are generally robust to word shuffling, so the small critical windows in our main experiments may not be specific to rationales. To test this, we apply the same \emph{n}-gram perturbation procedure directly to the \emph{question stem} instead of to the injected CoT rationale. In this setting, the recovery threshold shifts substantially upward: preserving ordinary question meaning requires much larger local windows than preserving the probe-time benefit of CoT rationales.

Figure~\ref{fig:wiki_qstem} compares these two controls. The Wikipedia control shows that generic topical text does not reproduce the benefit of rationale text. The question-stem control shows that low-window recovery is not simply a generic robustness property of language models under shuffling. Together, these results suggest that CoT-style rationales place useful probe-time signal in short local word neighborhoods, and that this pattern is specific to rationale context.

\section{Mechanistic Evidence for Local Co-occurrence}

We next examine where the difference between word-shuffled rationales and short local windows appears inside the probe model. First, we conduct layer-wise causal residual-stream patching on MMLU-Pro (Config~A). For each question and rationale, we run a word-shuffled (WS) forward pass and replace the residual states at all rationale positions with the corresponding states from the $n{=}3$ forward pass. We then measure how much of the target-answer margin gap between WS and $n{=}3$ is recovered.

As shown in Figure~\ref{fig:causal_patching}(a), patching Layers 4--12 recovers most of this margin gap, while the same intervention becomes much weaker in later layers. Position-specific patching further shows that the later positions inside each three-token block recover substantially more target-answer margin than the first position (Figure~\ref{fig:causal_patching}(b)). This pattern is consistent with the sharp behavioral recovery at $n{=}2$ and $n{=}3$: once a token is allowed to retain one or two immediate left neighbors, its representation can carry local contextual information that is absent from isolated word shuffling. These results suggest that the behavioral difference between WS and $n{=}3$ is largely formed in early-to-middle layers through locally conditioned token representations. Complete mechanistic results are provided in the appendix under \emph{Supplementary Mechanistic Generalization}.

Second, we test whether the short-window threshold is also reflected in the structure of the rationale text itself. We conduct a text-span analysis across MMLU-Pro, LogiQA, and MedQA. For each rationale, we remove explicit answer declarations, extract content words after stop-word filtering, and mark words that overlap with the question or the gold answer option. We then measure how much of this marked relevance mass falls inside contiguous spans of length at most $n$. In MMLU-Pro and MedQA, 64\%--70\% of this relevance mass lies within compact spans of length $\le 3$. By contrast, LogiQA shows more diffuse and longer relevant spans. As a model-based check, we also compute token-level gradient saliency on MMLU-Pro by differentiating the correct-answer log probability with respect to rationale-token embeddings; the resulting high-attribution spans have a short mean length of 2.22 tokens. This cross-dataset pattern mirrors the behavioral results: tasks with more compact answer-related spans also show faster recovery under short local windows.

Together, these two analyses support the local co-occurrence interpretation from different angles. Residual patching shows that the WS to $n{=}3$ difference is largely represented in early-to-middle layers, and text-span analysis shows that rationale evidence is often organized in localized spans. These results suggest that short-window recovery is not just an artifact of the perturbation procedure. Instead, CoT rationales often place useful probe-time evidence in short local neighborhoods, allowing the probe model to use these cues without reconstructing the full global reasoning chain.

\section{Generalization Across Task Formats}
\label{sec:gen}

Table~\ref{tab:cross_ngram} summarizes the generalization results. We use MMLU-Pro, LogiQA, and MedQA as the multiple-choice references, and include GSM8K \citep{cobbe2021gsm8k} and MATH-Hard \citep{hendrycksmath2021} as open-ended mathematical tasks. Due to space limits, the table reports results for three of the five datasets. Complete results for all five datasets are provided in the appendix under \emph{Experimental Details and Full Results}.

The same qualitative recovery pattern persists across task formats. Word-shuffled rationales remain well above IO, sentence-shuffled rationales stay close to full CoT, and short local windows recover much of the $\mathrm{WS}\rightarrow\mathrm{SS}$ gap. The critical window size, however, varies by task. On MMLU-Pro, all configurations reach half recovery by $n^\star \le 3$. On GSM8K and MATH-Hard, the threshold often shifts to $n^\star=5$, especially in Gemma~4 configurations. This is consistent with open-ended mathematical tasks, where useful evidence may require local combinations of quantities, operators, and intermediate expressions. Even so, recovery still occurs at windows far shorter than full rationale sentences or the original reasoning order. Thus, the local-window effect generalizes beyond multiple-choice answering, while the needed window size depends on the evidence required by the task.

\section{Conclusion}

We studied chain-of-thought prompting from a probe-time perspective by holding rationales fixed and systematically perturbing their structure. Across datasets and configurations, much of the probe-time benefit is recoverable from short local windows: word-shuffled rationales already outperform no-rationale baselines, and three-word windows recover over half of the $\mathrm{WS}\rightarrow\mathrm{SS}$ gap in most settings. Control experiments rule out explicit answer copying, simple lexical cues, generic topical context, and general robustness to shuffling as the main explanations. Mechanistic analyses further show that short-window gains are largely formed in early-to-middle layers, with answer-relevant evidence concentrated in local text spans. Together, these findings support a local co-occurrence activation (LCA) interpretation: probe-time CoT gains arise mainly from the rationale's word inventory and compact local neighborhoods, rather than from the global progression of the reasoning chain.

\bibliography{references}

\end{document}